\pdfoutput=1

\documentclass[11pt]{article}

\usepackage{EMNLP2022}

\usepackage{times}
\usepackage{latexsym}
\usepackage{microtype}
\usepackage{mathrsfs}
\usepackage{amsmath}
\usepackage{amssymb}
\usepackage{graphicx}
\usepackage{algorithm}
\usepackage{algpseudocode}
\usepackage{multirow}
\usepackage{booktabs}
\usepackage{url}
\usepackage{adjustbox}
\usepackage{enumitem}
\usepackage{lipsum}
\usepackage{xcolor,colortbl}

\usepackage{caption}
\usepackage{subcaption}

\setlist{topsep=1pt,itemsep=1pt,partopsep=1pt, parsep=1pt}

\usepackage[T1]{fontenc}

\usepackage[utf8]{inputenc}

\usepackage{microtype}

\usepackage{inconsolata}

\title{How Long Is Enough? Exploring the Optimal Intervals of \\ Long-Range Clinical Note Language Modeling}

\author{Samuel Cahyawijaya$^1$\thanks{\hspace{1mm} These authors contributed equally.}, Bryan Wilie$^{1*}$, Holy Lovenia$^{1*}$, \textbf{MingQian Zhong$^{2,3,4}$}, \\
\vspace{6pt} \textbf{Huan Zhong$^{2,3}$}, \textbf{Nancy Y. Ip$^{2,3,4}$}, \textbf{Pascale Fung$^1$} \\
$^1$Center for Artificial Intelligence Research (CAiRE), Department of Electronic and Computer Engineering, \\
The Hong Kong University of Science and Technology, Hong Kong, China \\ 
\vspace{6pt} \texttt{\{scahyawijaya, bwilie, hlovenia, pascale\}@ust.hk}\\
$^2$Division of Life Science, State Key Laboratory of Molecular Neuroscience, Molecular Neuroscience Center, \\
The Hong Kong University of Science and Technology, Clear Water Bay, Kowloon, Hong Kong, China \\
\vspace{6pt} \texttt{\{mzhongac,dorothyzhong,boip\}@ust.hk}\\
$^3$Hong Kong Center for Neurodegenerative Diseases, Hong Kong Science Park, Hong Kong, China \\
\vspace{6pt} \texttt{\{mzhongac,dorothyzhong,boip\}@ust.hk}\\
}

\begin{document}
\maketitle
\begin{abstract}

Large pre-trained language models (LMs) have been widely adopted in biomedical and clinical domains, introducing many powerful LMs such as bio-lm and BioELECTRA. However, the applicability of these methods to real clinical use cases is hindered, due to the limitation of pre-trained LMs in processing long textual data with thousands of words, which is a common length for a clinical note. In this work, we explore long-range adaptation from such LMs with Longformer, allowing the LMs to capture longer clinical notes context. We conduct experiments on three n2c2 challenges datasets and a longitudinal clinical dataset from Hong Kong Hospital Authority electronic health record (EHR) system to show the effectiveness and generalizability of this concept, achieving 10\% F1-score improvement. Based on our experiments, we conclude that capturing a longer clinical note interval is beneficial to the model performance, but there are different cut-off intervals to achieve the optimal performance for different target variables. Our code is available at~\url{https://github.com/HLTCHKUST/long-biomedical-model}.


\end{abstract}

\section{Introduction}

Clinical note is one of the most abundant data available in EHR systems, which records most of the patient interaction with the hospital services, such as consultation with doctors, procedure note, laboratory report, discharge summary, etc.\footnote{\url{https://www.healthit.gov/isa/uscdi-data-class/clinical-notes}} Despite retaining rich clinical information, clinical notes are highly unstructured and composed of non-standardized information, which curbs the potential practicality of such information. Large pre-trained LMs, such as BERT~\cite{devlin2019bert}, RoBERTa~\cite{liu2020roberta}, GPT-2~\cite{radford2019language}, etc., have been shown to work well in extracting crucial information from clinical notes by utilizing transfer learning and attention mechanism~\cite{ji2021does,alsentzer2019clinicalbert,lewis-etal-2020-pretrained}. The adaptation of these models to biomedical and clinical domain emphasizes this success, establishing many new state-of-the-art performances on multiple biomedical and clinical benchmarks~\cite{peng2019blue,gu2021blurb,zhang-etal-2022-cblue}.

While the attention mechanism embedded in the pre-trained models enables them to achieve great performance, it is to be noted that it also causes a quadratic growth in computation cost with respect to input sequence length~\cite{tay2022efficient,wang2020linformer,cahyawijaya2022snp2vec}. This makes efficiently processing long documents with pre-trained LMs difficult, especially in clinical note modeling, in which a single clinical note tends to consist of hundreds or even thousands of words~\cite{uzuner2008smoking,uzuner2009obesity,stubbs2015dg,gehrmann2018clinical,johnson2019mimiciii,stubbs2019hz}. Current approaches to this problem commonly involve truncation, chunking, or windowing of the long input sequence, preventing the models from acquiring an entire medical record information provided by a whole clinical note. Considering that clinical note modeling requires capturing and understanding the underlying long-term dependencies in the clinical notes, this certainly puts a limit on their predictive capability. For this reason, to maximize the models' capability without sacrificing a part of the input clinical notes, we explore the application of long-range adaptation through linear attention mechanism~\cite{dai2019trfxl,beltagy2020longformer,wang2020linformer,choromanski2021rethinking}, which reduces the computation cost of attention from quadratic to linear in regards to input sequence length.

In this work, we focus on assessing the benefit of capturing longer clinical notes on large pre-trained LMs to n2c2 (National Clinical NLP Challenges)\footnote{\url{https://n2c2.dbmi.hms.harvard.edu/}} clinical tasks 
by adapting a linear attention mechanism, i.e., Longformer~\cite{beltagy2020longformer}. 
Furthermore, to test the generality of this approach,
we evaluate it on a longitudinal clinical note corpus from Hong Kong Hospital Authority EHR system, which covers records from 43 hospitals in Hong Kong. Lastly, we hypothesize that modeling longer interval of clinical notes improves the prediction quality of the models on any clinical task. To prove our hypothesis, we conduct our experiment using different context-length, allowing the model to access various intervals of clinical notes. Our result suggests that a longer interval of clinical notes increases the prediction quality of the models in most cases, but there is a limit of context length required depending on the target variable.

Our contributions in this work can be summarized in three-fold: 
\begin{itemize}
    \item  We assess the effectiveness of capturing longer interval of clinical notes on biomedical and clinical large pre-trained LMs on three n2c2 challenges which increase the performance by $\sim$10\% F1-score,
    \item We evaluate the generalization of this approach using longitudinal clinical note data gathered in Hong Kong Hospital Authority EHR system on two clinical tasks, i.e., disease risk and mortality risk predictions, which improve the performance by $\sim$5-10 F1-score, 
    \item We observe that each target variable has a different optimal clinical notes cut-off interval and we conclude that the optimal cut-off interval for mortality risk prediction is $\sim$2-3 months, while for disease risk prediction, it requires 3.5 years or even longer interval to achieve the optimal performance.
\end{itemize}

\section{Related Works}


\paragraph{Clinical Note Modeling}
Clinical notes have been utilized for various applications in healthcare. Text mining methods for analyzing pharmacovigilance signals using clinical notes have been explored and yield promising results~\cite{haerian2012pharmacovigilance,LePendu2012AnnotationAF,lependu2013phrmacovigilance}. Clinical notes with other EHR data are also employed for estimating the readmission time and mortality risk of the next patient encounter~\cite{diabeticsreadmmission,rajkomar2018scalabel}.
Clinical note data is also effective for analyzing disease comorbidity, such as mental illness~\cite{wu2013comorbid}, autoimmune diseases~\cite{escudie2017comorbid}, and obesity~\cite{pantalonee2017comorbid}. Predicting disease risk using clinical note data has also been explored~\cite{miotto2016deeppatient,choi2018mime,liu2019pharmacovigilance,liu2018deep,Koleck_Dreisbach_Bourne_Bakken_2019}. Despite all the efforts in clinical note modeling, to the best of our knowledge, how clinical note interval impacts the performance of pre-trained LMs has never been studied.

\paragraph{Biomedical and Clinical Pre-trained LMs}

Self-supervised pre-training LMs employing transformer-based architectures~\cite{vaswani2017transformer}, such as BERT~\cite{devlin2019bert}, RoBERTa~\cite{liu2020roberta}, and ELECTRA~\cite{clark2019electra}, have thrived in various general domain NLP benchmarks~\cite{wang2018glue, rajpurkar2016squad,ladhak2020wikilingua,lai2017race,wilie2020indonlu,cahyawijaya2021indonlg,park2021klue}. To extend the understanding of these LMs to the linguistic properties in biomedical and clinical domain, a generation of LMs exploiting biomedical and clinical corpora emerges.

In 2019,~\citet{alsentzer2019clinicalbert} introduce BioBERT, an extended version of BERT pre-trained on large-scale biomedical data (i.e., PubMed abstracts and PMC full-text articles) which surpasses off-the-shelf BERT
in three fundamental downstream tasks in biomedical domain.
Due to the linguistic differences exhibited by non-clinical biomedical texts and clinical texts,~\citet{alsentzer2019clinicalbert} introduce ClinicalBERT by fine-tuning BERT and BioBERT on the MIMIC-III corpus, and improve the performance over five clinical NLP tasks.

Unlike prior works, PubMedBERT~\cite{pubmedbert} performs biomedical pre-training from scratch, which offers larger performance gains over various biomedical downstream tasks in the BLURB benchmark. 
Similarly, bio-lm~\cite{lewis-etal-2020-pretrained} employs recent pre-training advances, utilizes various biomedical and clinical corpora for pre-training, and achieves the highest performance on 
9 biomedical and clinical NLP tasks. In 2021, BioELECTRA~\cite{kanakarajan-etal-2021-bioelectra}, a general domain ELECTRA~\cite{clark2019electra} pre-trained on biomedical corpora, sets the new state-of-the-art performance for all datasets in the BLURB benchmark and
4 datasets in the BLUE benchmark~\cite{peng2019blue}.




\paragraph{Long Sequence Language Modeling}
Recent progress in language modeling is dominated by transformer-based models which shows a remarkable results on numerous tasks. Nevertheless, these models have limited capability to process long-range clinical notes data due to its quadratic attention complexity.
Various approaches have been introduced to reduce this complexity problem, such as recurrence approach~\cite{dai2019trfxl,rae2020compressive}, sparse and local attention patterns~\cite{kitaev2020reformer,qiu-etal-2020-blockwise,child2019generating,zaheer2020big,beltagy2020longformer}, low-rank approximation~\cite{wang2020linformer,winata2020lrt}, and kernel methods~\cite{katharopoulos2020lineartrf, choromanski2021rethinking}. Adaptation from existing pre-trained models to some of these methods have also been explored and show the potential for knowledge transfer~\cite{beltagy2020longformer, choromanski2021rethinking}. In this work, we utilize Longformer~\cite{beltagy2020longformer} to enable the model to capture long-range clinical note information.

\section{Methodology}
\label{sec:methodology}

\subsection{Problem Definition}

Clinical notes are narrative patient data relevant to the context identified by note types\footnote{\url{https://www.healthit.gov/isa/uscdi-data-class/clinical-notes}}. There are multiple types of clinical notes, e.g., discharge summary, consultation note, progress note, lab report, etc. In general, a single clinical note consists of a text narrative and additional metadata defining the clinical note, e.g., note identifier, recording timestamp of the note, etc. In n2c2 challenges, a single clinical note is presented in a textual format with the metadata written on top of the text narrative, while a longitudinal clinical note is presented as a concatenation of several clinical notes with a separator text placed between two clinical notes. This clinical note is usually long, ranging from several hundreds to thousands words, while most existing biomedical and clinical pre-trained LMs can only capture up to 512 subwords, which is insufficient to capture the whole content of most clinical notes.

\subsection{Long-Range Clinical Note LMs}
\label{sec:long-range-methodology}

We increase the capacity of LMs to process longer clinical notes by adapting Longformer~\cite{beltagy2020longformer} to the existing biomedical and clinical pre-trained LMs. Longformer enables linear attention mechanism by dividing single quadratic all-to-all attention into two attention steps, i.e., sliding-window and global attentions. Sliding-window attention allows each token to attend to neighboring tokens, while global attention allows some, usually a few, tokens to attend to all tokens, hence has a better computation complexity compared to the quadratic attention mechanism. It is to be noted that when extending an original transformer-based model into a Longformer, some new parameters are introduced, i.e., the new positional embeddings, the sliding-window projection parameters, and global attention projection parameters. For the positional embeddings, following~\cite{beltagy2020longformer}, we copy the weights of the pre-trained positional embeddings to initialize the new positional embeddings. For the sliding-window and global attention parameters, we initialize both projection parameters with the pre-trained projection parameters.

\begin{table*}[t!]
    \centering
    \resizebox{1.0\linewidth}{!}{
        \begin{tabular}{lcccccccccc}
        \toprule
        \multirow{2}{*}{\vspace{-1mm}\bfseries{Dataset}} & \multirow{2}{*}{\vspace{-1mm}\bfseries{|Train|}} & \multirow{2}{*}{\vspace{-1mm}\bfseries{|Test|}} & \multicolumn{4}{c}{\bfseries{Word count}} &
        \multirow{2}{*}{\vspace{-1mm}\bfseries{{Longitudinal?}}} &
        \multirow{2}{*}{\vspace{-1mm}\bfseries{\#Label}} & \multirow{2}{*}{\vspace{-1mm}\bfseries{\#Class}} \\
        \cmidrule{4-7}
         & & & \textbf{Median} & \textbf{Q3} & \textbf{95\%} &   \textbf{Max} & 
         & & & \\
        \midrule
        \textbf{2006 Smoking} & 398 & 104 & 677 & 1096 & 1775 & 3023 & No & 5 & 1 \\
        \textbf{2008 Obesity (Textual)} & 730 & 507 & 1084 & 1425 & 2094 & 4280 & No & 16 & 4 \\
        \textbf{2008 Obesity (Intuitive)} & 730 & 507 & 1084 & 1425 & 2094 & 4280 & No & 16 & 4 \\
        \textbf{2018 Cohort Selection} & 202 & 86 & 2550 & 3235 & 4578 & 7070 & Yes & 13 & 1 \\
        \bottomrule
        \end{tabular}}
    \caption{The overall statistics of the n2c2 datasets used in our experiment.}
    \label{tab:n2c2-datasets}
\end{table*}

\section{Long-Range Clinical Note LMs on n2c2 Challenges}

We assess the effectiveness of long-range clinical note LMs on US-based clinical note datasets from three n2c2 challenges. Additionally, we also evaluate six different pre-trained LMs without long-range adaptation to benchmark the performance of the biomedical and clinical LMs.



\subsection{Dataset}
\label{sec:n2c2-dataset}

We use three clinical datasets concentrating on classifying diverse clinical problems from n2c2. These datasets are: 1) n2c2 2006 smoking challenge, focusing on predicting smoking status of patients based on their discharge summary; 2) n2c2 2008 obesity challenge, focusing on recognizing obesity and its comorbidities of patients through their discharge summary; and 3) n2c2 2018 cohort selection challenge, focusing on determining if a patient meets selection criteria of certain clinical trials cohorts through longitudinal clinical notes. We utilize BigBIO framework~\cite{fries2022bigbio}\footnote{\url{https://github.com/bigscience-workshop/biomedical}} to load the n2c2 datasets. We provide overview of these datasets in Table \ref{tab:n2c2-datasets}. 

\paragraph{n2c2 2006 Smoking Challenge}
We utilize the smoking prediction subtask from n2c2 2006 challenge~\cite{uzuner2008smoking}, where each data instance consists of a de-identified discharge summary annotated by two pulmonologists with smoking status. This smoking status can be either past smoker" (when it is explicitly stated that the patient is a past smoker or that the patient used to smoke but has stopped for at least a year), "current smoker" (when it is explicitly stated that the patient is a current smoker or that the patient has smoked within the past year), "smoker" (when there is not enough temporal information to classify whether a patient is a "past smoker" or "current smoker"), "non-smoker" (when a patient's discharge summary indicates an absence of smoking habit), or "unknown" (when there is no mention of smoking). 

\begin{table*}[t]
    \centering
    \resizebox{1.0\linewidth}{!}{
        \begin{tabular}{ccccccccc}
        \toprule
        \multirow{2}{*}{} & \multicolumn{2}{c}{\textbf{2006 Smoking}} & \multicolumn{2}{c}{\textbf{2008 Obesity (Text.)}} & \multicolumn{2}{c}{\textbf{2008 Obesity (Intui.)}} & \multicolumn{2}{c}{\textbf{2018 Cohort Selection}} \\
        \cmidrule(lr){2-3} \cmidrule(lr){4-5} \cmidrule(lr){6-7} \cmidrule(lr){8-9}
         & \textbf{micro-f1} & \textbf{macro-f1} & \textbf{micro-f1} & \textbf{macro-f1} & \textbf{micro-f1} & \textbf{macro-f1} & \textbf{micro-f1} & \textbf{macro-f1} \\
        \toprule
        
        \multicolumn{9}{>{\columncolor[gray]{.9}}c}{\textbf{\textit{Baseline}}} \\
        \textbf{Top-5 scorer} & 88.00\% & 69.00\% & 97.04\% & 77.18\% & 95.58\% & 63.44\% & 90.30\% & \\
        \textbf{Top-10 scorer} & 86.00\% & 58.00\% & 96.39\% & 61.40\% & 95.08\% & 62.87\% & 87.70\% & \\
        \midrule
        
        \multicolumn{9}{>{\columncolor[gray]{.9}}c}{\textbf{\textit{Pre-trained Language Model}}} \\
        \textbf{BERT-cased} & 61.63\% & 31.79\% & 82.47\% & 38.73\% & 81.69\% & 51.71\% & 72.80\% & 48.45\% \\
        \textbf{BERT-uncased} & 65.63\% & 41.12\% & 85.73\% & 40.83\% & 83.46\% & 53.28\% & \underline{74.86\%} & 51.32\% \\
        \textbf{clinicalBERT} & 56.59\% & 39.34\% & 85.64\% & 40.64\% & 85.20\% & 54.88\% & 72.83\% & \underline{49.99\%} \\
        \textbf{PubMedBERT} & 69.38\% & 41.65\% & \textbf{88.98\%} & \underline{46.27\%} & \textbf{87.11\%} & \textbf{56.47\%} & 74.78\% & 49.94\% \\
        \textbf{bio-lm} & \textbf{71.44\%} & \textbf{49.43\%} & 86.57\% & 43.15\% & 84.92\% & 54.73\% & \textbf{75.03\%} & \textbf{52.18\%} \\
        \textbf{BioELECTRA} & \underline{70.72\%} & \underline{48.26\%} & \underline{86.71\%} & \textbf{48.26\%} & \underline{85.31\%} & \underline{55.00\%} & 74.32\% & 49.10\% \\
        \midrule

        \multicolumn{9}{>{\columncolor[gray]{.9}}c}{\textbf{\textit{Long-range Pre-trained Language Model}}} \\
        \textbf{bio-lm (1024)} & 82.12\% & 55.72\% & 92.52\% & 50.36\% & 90.36\% & 59.13\% & 77.03\% & 53.94\% \\
        \textbf{bio-lm (2048)} & \textbf{86.01\%} & 62.30\% & 96.44\% & 55.99\% & \underline{94.76\%} & 62.61\% & 76.76\% & 52.93\% \\
        \textbf{bio-lm (4096)} & 84.52\% & 57.76\% & \textbf{97.11\%} & 55.68\% & \textbf{95.48\%} & \underline{63.19\%} & 79.42\% & 57.85\% \\
        \textbf{bio-lm (8192)} & 84.66\% & 59.49\% & \underline{97.07\%} & 55.08\% & \textbf{95.48\%} & \textbf{63.20\%} & \underline{81.43\%} & \underline{\textbf{61.95\%}} \\
        
        \textbf{BioELECTRA (1024)} & 82.98\% & \underline{63.35\%} & 93.54\% & 54.47\% & 90.40\% & 59.12\% & 74.95\% & 51.69\% \\
        \textbf{BioELECTRA (2048)} & 82.84\% & 61.09\% & 96.03\% & \underline{56.08\%} & 91.69\% & 60.21\% & 77.59\% & 54.39\% \\
        \textbf{BioELECTRA (4096)} & 80.40\% & 57.22\% & 95.81\% & 56.06\% & 92.88\% & 61.12\% & 79.10\% & 56.38\% \\
        \textbf{BioELECTRA (8192)} & \underline{85.21\%} & \textbf{64.32\%} & 96.20\% & \textbf{59.59\%} & 92.78\% & 61.09\% & \textbf{81.63\%} & \underline{58.44\%} \\
        \bottomrule
        \end{tabular}}
    \caption{Evaluation results of our experiments on the n2c2 datasets. Top-5 and Top-10 scorers are retrieved from the submission benchmark of corresponding challenge. The number inside the bracket denotes the length of  context that can be captured by the model. \textbf{Bold} and \underline{underline} denotes the first and second best scores within a group.}
    \label{tab:n2c2-results}
\end{table*}

\paragraph{n2c2 2008 Obesity Challenge}
The n2c2 2008 obesity challenge~\cite{uzuner2009obesity} consists of 1027 pairs of de-identified discharge summaries and 16 disease labels. The disease labels include obesity and its 15 comorbidities, e.g., asthma, atherosclerotic cardiovascular disease (CAD), congestive heart failure (CHF), depression, diabetes mellitus (DM), gallstones/cholecystectomy, gastroesophageal reflux disease (GERD), gout, hypercholesterolemia, hypertension (HTN), hypertriglyceridemia, obstructive sleep apnea (OSA), osteoarthritis (OA), peripheral vascular disease (PVD), and venous insufficiency.

The annotation for each discharge summary is done by providing each disease label with either "present", "absent", "questionable", or "unmentioned". 
The dataset has two types of annotations, i.e., textual judgement (only based on related explicit statements) and intuitive judgement (based on everything written in the discharge summaries).
We use both annotations in our experiments and report the evaluation scores for each annotation.

\paragraph{n2c2 2018 Cohort Selection Challenge}

The 2018 Shared Task 1: Clinical Trial Cohort Selection~\cite{stubbs2019hz} reuses 288 patient records from the 2014 n2c2 shared task dataset~\cite{stubbs2015dg} and reframes it as a cohort selection task, which requires an automatic evaluation of whether a patient fits or does not fit in certain cohorts according to their longitudinal de-identified clinical notes, ranging between 2-5 clinical notes.

The cohorts or selection criteria used in the dataset as labels are: DRUG-ABUSE (current or past usage of drugs), ALCOHOL-ABUSE (current alcohol intake over weekly recommended limit), ENGLISH (English-speaking patient), MAKES-DECISIONS (patients required to make their own medical decisions), ABDOMINAL (history of related surgery), MAJOR-DIABETES (major diabetes-related complication), ADVANCED-CAD (advanced cardiovascular disease), MI-6MOS (myocardial infarction in the past 6 months), KETO-1YR (diagnosis of ketoacidosis in the past year), DIETSUPP-2MOS (dietary supplement intake in the past 2 months, excluding vitamin D), ASP-FOR-MI (usage of aspirin to prevent MI), HBA1C (any hemoglobin A1c value between 6.5\% and 9.5\%), and CREATININE (serum creatinine above the upper limit of normal). Two annotators with medical expertise classify each label of a patient's set of clinical notes as either "met" or "not met".

\subsection{Models}

In this experiment, we compare several pre-trained LMs, covering two variants of BERT model representing general domain LMs, i.e., uncased BERT\footnote{\url{https://huggingface.co/bert-base-uncased}} and cased BERT\footnote{\url{https://huggingface.co/bert-base-cased}}, two variants of biomedical domain LMs, i.e. PubMedBERT~\cite{gu2021blurb}\footnote{\url{https://huggingface.co/microsoft/BiomedNLP-PubMedBERT-base-uncased-abstract}} and BioELECTRA~\cite{kanakarajan-etal-2021-bioelectra}\footnote{\url{https://huggingface.co/kamalkraj/bioelectra-base-discriminator-pubmed}}, one variant of clinical domain LM, i.e., ClinicalBERT~\cite{alsentzer2019clinicalbert}\footnote{\url{https://huggingface.co/emilyalsentzer/Bio_ClinicalBERT}}, and one variant of mixed biomedical and clinical domains LM, i.e., bio-lm~\cite{lewis-etal-2020-pretrained}\footnote{\url{https://huggingface.co/EMBO/bio-lm}}.

To enable longer context clinical note modeling, we adapt Longformer~\cite{beltagy2020longformer} with the initialization strategy specified in \S\ref{sec:long-range-methodology}. We conduct experiments with four different context lengths, i.e., $\{1024, 2048, 4096, 8192\}$ on two pre-trained LMs variants, i.e., BioELECTRA and bio-lm. 

\subsection{Training and Evaluation}
\label{sec:n2c2-exp}

Following BERT, RoBERTa, and bio-lm experiments, we tune the learning rate for all BERT and RoBERTa models from [1e-5, 2e-5, 3e-5]. While for the BioELECTRA model, following ELECTRA~\cite{clark2019electra} and BioELECTRA~\cite{kanakarajan-etal-2021-bioelectra}, we tune the learning rate from [5e-5, 1e-4, 2e-4]. In all experiments, we use a batch size of 8, and a linear learning rate decay. For the n2c2 2006 and n2c2 2008 tasks, we train the models for 50 epochs, while for the n2c2 2018 task, we train the models for 80 epochs. For the evaluation, we incorporate the official evaluation metrics defined for each challenge. All of them report micro-F1 and macro-F1 scores. 

\begin{figure*}[t]
\centering
\begin{subfigure}{.46\textwidth}
  \centering
  \includegraphics[width=1.0\linewidth]{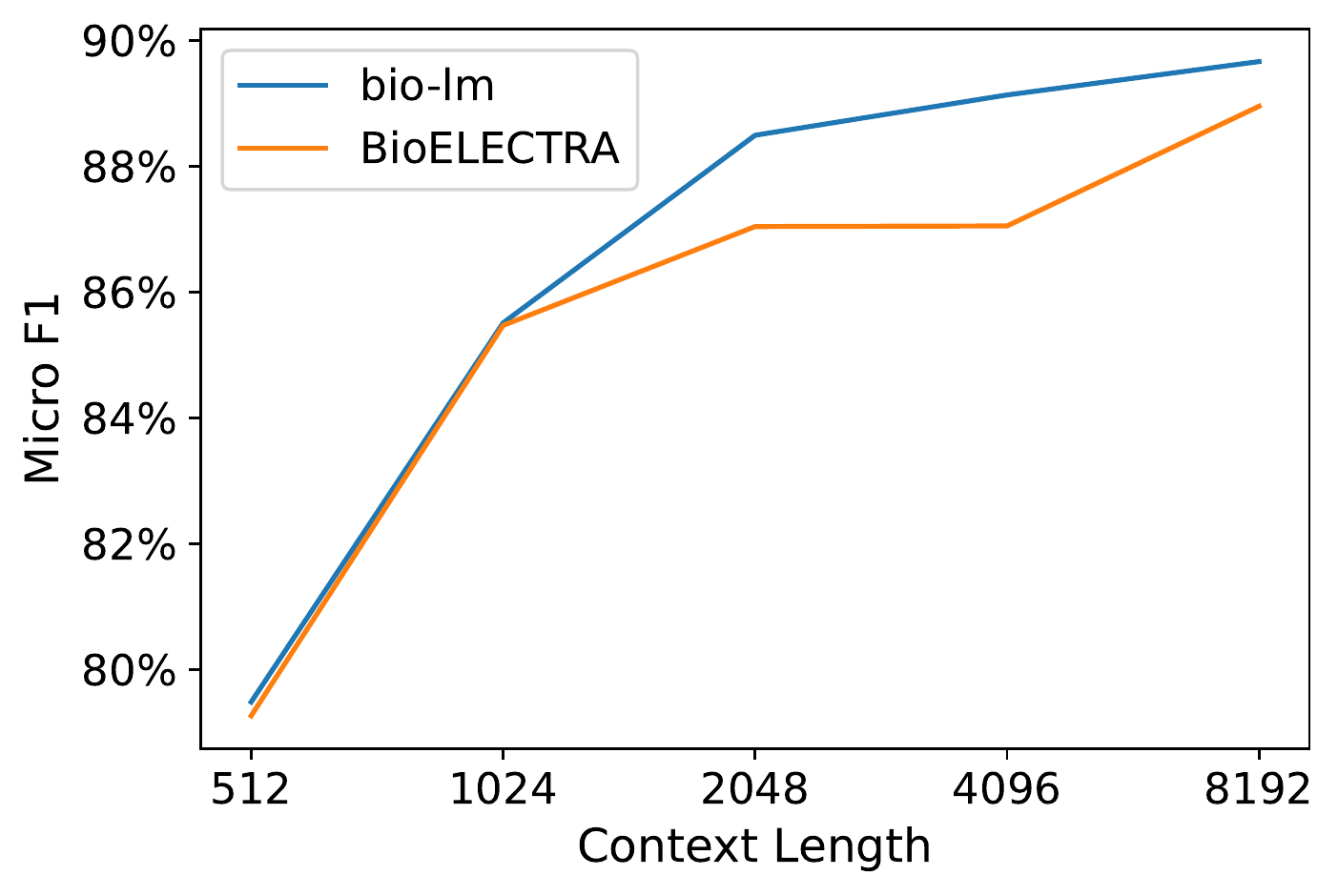}
\end{subfigure}%
\hspace{10pt}
\begin{subfigure}{.46\textwidth}
  \centering
  \includegraphics[width=1.0\linewidth]{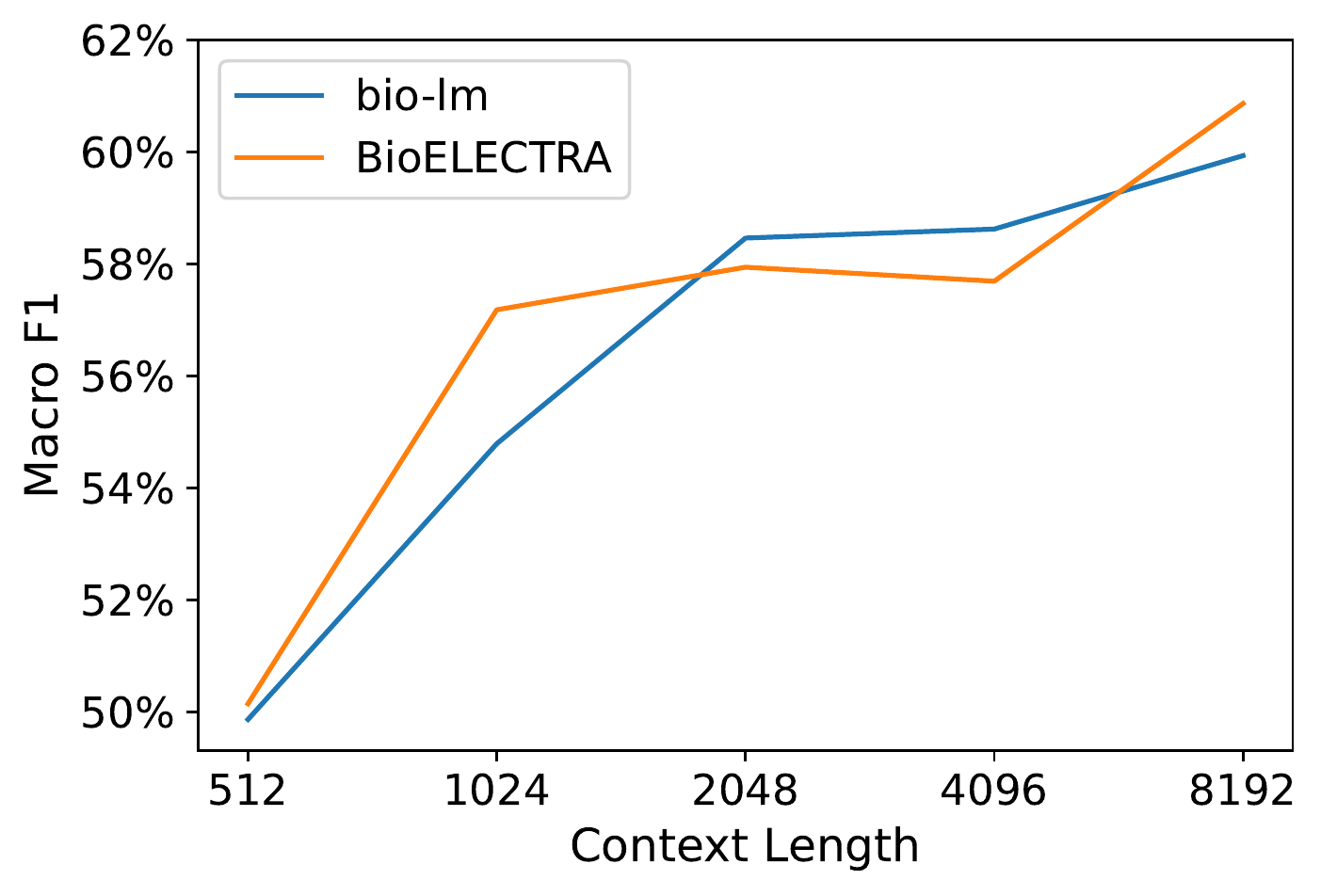}
\end{subfigure}
\caption{Effect of capturing longer clinical notes context to the evaluation performance, i.e., on micro-F1 \textbf{(Left)} and macro-F1 \textbf{(Right)}, averaged over the context length across the evaluated n2c2 tasks.}
\label{fig:effect-long-model}
\end{figure*}

\subsection{Result and Analysis}

As shown in Table~\ref{tab:n2c2-results}, in general, domain-specific LMs yield higher performance compared to general domain LMs, except for ClinicalBERT which performs on a par with the general domain BERT models. PubMedBERT, bio-lm, and BioELECTRA produce comparable evaluation performances across all tasks, with $\sim$2-5\% higher F1-score compared to the general domain BERT and ClinicalBERT. Nevertheless, the scores are much lower compared to the Top-10 scorer on the challenge benchmark since the models can only capture partial information of the clinical note data. 

By increasing the context length of the model, the performance rises significantly. Comparing with the original pre-trained versions of the models, the best performing long-range pre-trained LM improves the evaluation performance by $\sim$10\% F1-score in all datasets. As shown in Figure~\ref{fig:effect-long-model}, models with longer context length tend to perform better, but the performance gain is limited to the length of the clinical notes in the dataset. For instance, on the n2c2 2006 dataset, the performance improvement of both bio-lm and BioELECTRA models are steeper from context length 512 to 1024 rather than from context length 1024 to 2048, 2048 to 4096, and 4096 to 8192. This is because a huge portion of the notes in the datasets can be sufficiently captured within 1024 subwords. In contrast, the performance improvement on the n2c2 2018 dataset is more linear per context length step since most of the length of the clinical notes is much longer than the other two datasets. Every step of extending the context length provides more information to the model, which is likely to improve the model performance considerably. 

On the n2c2 2006 and 2008 challenges, our best performing models mostly achieve a comparable score to the Top-10 or Top-5 scorer of the corresponding challenge benchmark. This is a remarkable feat since our models neither utilize any ensemble method, incorporate any clinical expert, nor exploit external data--common practices used by the top scorers in the challenge benchmarks.


\section{Long-Range Clinical Note LMs on Hong Kong Longitudinal Dataset}

We assess the generalization and effectiveness of long-range clinical notes LMs on Hong Kong longitudinal clinical note data. We construct a longitudinal dataset with two target variables, i.e., disease risk and mortality risk, and evaluate long-range LMs on the dataset. In addition, we add a baseline model, which takes high-level features extracted from the corresponding tabular data provided by the EHR system as the input, to assess the effectiveness of clinical note modeling.

\subsection{Dataset Construction}

We construct a longitudinal clinical note dataset for disease risk and mortality risk predictions from anonymized cancer cohort patient records gathered in the Hong Kong Hospital Authority EHR system covering 43 hospitals in Hong Kong. The patient records span across the year 2000 and 2018. We exclude all patients having less than two clinical notes and gather a total of $\sim$300,000 patients. To construct labelled data for the supervised learning, from patient $P_i$ with $T$ clinical records, we build $T-1$ labelled autoregressive data $\mathcal{D}^{Pi} = \{\{C_k^{P_i}\}_{k=1}^t, Y^{P_i}_{t+1}\}_{t=1}^{T-1}$, where $\{C_k^{P_i}\}_{k=1}^t$ denotes $t$ prior clinical notes of the patient $P_i$, and $Y_{t+1}^{P_i}$ denotes the target criterion retrieved from the ${t+1}^{th}$ clinical record of the patient $P_i$. We collect over $\sim$2M labelled clinical notes from all patients with two targets: disease risk and mortality risk.

\begin{table}[t]
	    \resizebox{1.0\linewidth}{!}{
        \begin{tabular}{@{}ccccc@{}}
        \toprule
           \multirow{2}{*}{\textbf{Split}} & \multirow{2}{*}{\textbf{\# Patients}} & \textbf{\# Seen patient} & \textbf{\# Unseen patient} \\      
             &  & \textbf{records} & \textbf{records} \\
        \midrule
            Train & 278,253 & 2,027,561 & - \\
            Valid & 3,621 & 3,177 & - \\
            Test & 17,903 & 15,541 & 2,362 \\   \midrule
            Total & 299,777 & 2,046,279 & 2,362 \\
        \bottomrule
        \end{tabular}
    }
	\caption{The overall statistics of our Hong Kong longitudinal dataset. \textbf{\# Seen patient records} and \textbf{\# Unseen patient records} indicate the number of records on the \textit{seen} and \textit{unseen} test set respectively.}
	\label{tab:dataset-statistics}
\end{table}

\begin{table*}[!t]
    \centering
    \resizebox{0.85\linewidth}{!}{
    \begin{tabular}{llcccccc}
    \toprule
        \multirow{3}{*}{\textbf{Test set}} & \multirow{3}{*}{\textbf{Models}} & \multicolumn{4}{c}{\textbf{Diagnosis}} & \multicolumn{2}{c}{\textbf{Mortality}} \\ \cmidrule(lr){3-6} \cmidrule(lr){7-8} 
        & & \textbf{Top-1} &  \textbf{Top-3} &    \textbf{Top-5} & \textbf{F1} & \textbf{F1} & \textbf{AUC} \\
    \midrule
        \multirow{4}{*}{\textit{Seen}} & \textbf{EHR-FFN} & 64.3\% & 75.7\% & 80.3\% & 40.6\% & 49.5\% & 78.1\% \\
        & \textbf{BioELECTRA (512)} & 76.2\% & 88.6\% & 91.8\% & 51.6\% & 61.5\% & \textbf{92.0\%} \\
        & \textbf{BioELECTRA (2048)} & \underline{79.8\%} & \underline{91.5\%} & \underline{94.3\%} & \underline{54.2\%} & \textbf{65.3\%} & \underline{91.9\%} \\
        & \textbf{BioELECTRA (8192)} & \textbf{81.3\%} & \textbf{92.9\%} & \textbf{95.5\%} & \textbf{55.7\%} & \underline{64.9\%} & 91.8\% \\
    \midrule
        \multirow{4}{*}{\textit{Unseen}} & \textbf{EHR-FFN} & 17.8\% & 32.9\% & 43.1\% & 9.5\% & 49.6\% & 73.9\% \\
        & \textbf{BioELECTRA (512)} & 63.4\% & 78.6\% & 83.7\% & 43.1\% & \underline{52.2\%} & 84.8\% \\
        & \textbf{BioELECTRA (2048)} & \underline{66.3\%} & \underline{81.2\%} & \underline{85.9\%} & \underline{45.1\%} & 52.0\% & \underline{85.8\%} \\
        & \textbf{BioELECTRA (8192)} & \textbf{69.1\%} & \textbf{84.0\%} & \textbf{88.2\%} & \textbf{46.8\%} & \textbf{52.3\%} & \textbf{88.1\%} \\
    \bottomrule
    \end{tabular}
    }
    \caption{Evaluation results of our experiments on the \textit{seen} patient test set and the \textit{unseen} patient test set. \textbf{Bold} and \underline{underline} denotes the first and the second best score on each test set, respectively.}
    \label{tab:exp-result}
\end{table*}

We take the last two health records from all patient records in the year 2018 as the validation and test sets. To assess the generalization to new patient data, we omit some patient data from the training set and only used the last labelled record of those patients as the \textit{unseen} test set. The remaining test data becomes the \textit{seen} test set. The dataset statistics is shown in Table~\ref{tab:dataset-statistics}. For the disease risk estimation, we take the final disease diagnosis on the next clinical record as the label. For cancer diseases, we group the diagnosis based on the cancer site categorization from the Hong Kong Cancer Registry\footnote{\url{https://www3.ha.org.hk/cancereg/allages.asp}}, while for other diseases, we take the first three digits of the ICD-10 codes. In total, there are 79 classes for disease risk estimation. For the mortality label, we retrieve the mortality status from the discharge code from the next clinical record of the corresponding patient. The label distribution of the dataset is shown in Appendix~\ref{app:label-dist}.

\subsection{Models}

We experiment using Longformer with three variants of sequence length, i.e., $\{512, 2048, 8192\}$. We initialized all models with the same pre-trained BioELECTRA~\cite{kanakarajan-etal-2021-bioelectra} checkpoint as in \S\ref{sec:n2c2-exp}. 
To assess the effectiveness of clinical note modeling, we employ another baseline using a 4-layer feedforward model ($\sim$5M parameters), which takes an input of 3,942 dimension high-level features from the EHR database (EHR-FFN). Similar to DeepPatient~\cite{miotto2016deeppatient}, we extract high-level features from the diagnoses, medications, procedures, and laboratory test records by counting the occurrence of each feature type. In addition, we also add other features such as length of stay, the indicator for emergency unit admission, age group, etc. The details of EHR-FFN and the extracted features are shown in Appendix~\ref{app:tabular-models}.

\subsection{Training and Evaluation}


We train all of the models with an initial learning rate of 5e-5, batch size of 48, and a linear learning rate decay. We train the model for 3 epochs and test the model with the best validation score.
For evaluating the diagnosis label, we incorporate the F1-score along with the Top-1, Top-3,and Top-5 accuracy scores. For the mortality label, we incorporate F1-score and AUC. The evaluation is conducted on two different test sets: (i) the \textit{seen} patient test set and (ii) the \textit{unseen} patient test set.



\subsection{Results and Analysis}


\paragraph{Effect of Clinical Note Modeling}

We show our experiment results for the \textit{seen} and the \textit{unseen} test sets in Table~\ref{tab:exp-result}. All BioELECTRA models yield higher results than the EHR-FFN for both test sets, showing the effectiveness of clinical note modeling for disease risk and mortality risk predictions using EHR data. From the comparison between different clinical notes interval of the BioELECTRA model, we found that modeling longer clinical note interval will likely increase the performance on both tasks. This behavior aligns with the results reported in \S\ref{tab:n2c2-results}.  Nevertheless, this behavior does not apply to the mortality risk prediction on the \textit{seen} test set. We describe this phenomenon further in \S\ref{sec:optimal}.


\begin{figure*}[t]
\centering
\begin{subfigure}{.48\textwidth}
  \centering
  \includegraphics[width=\linewidth]{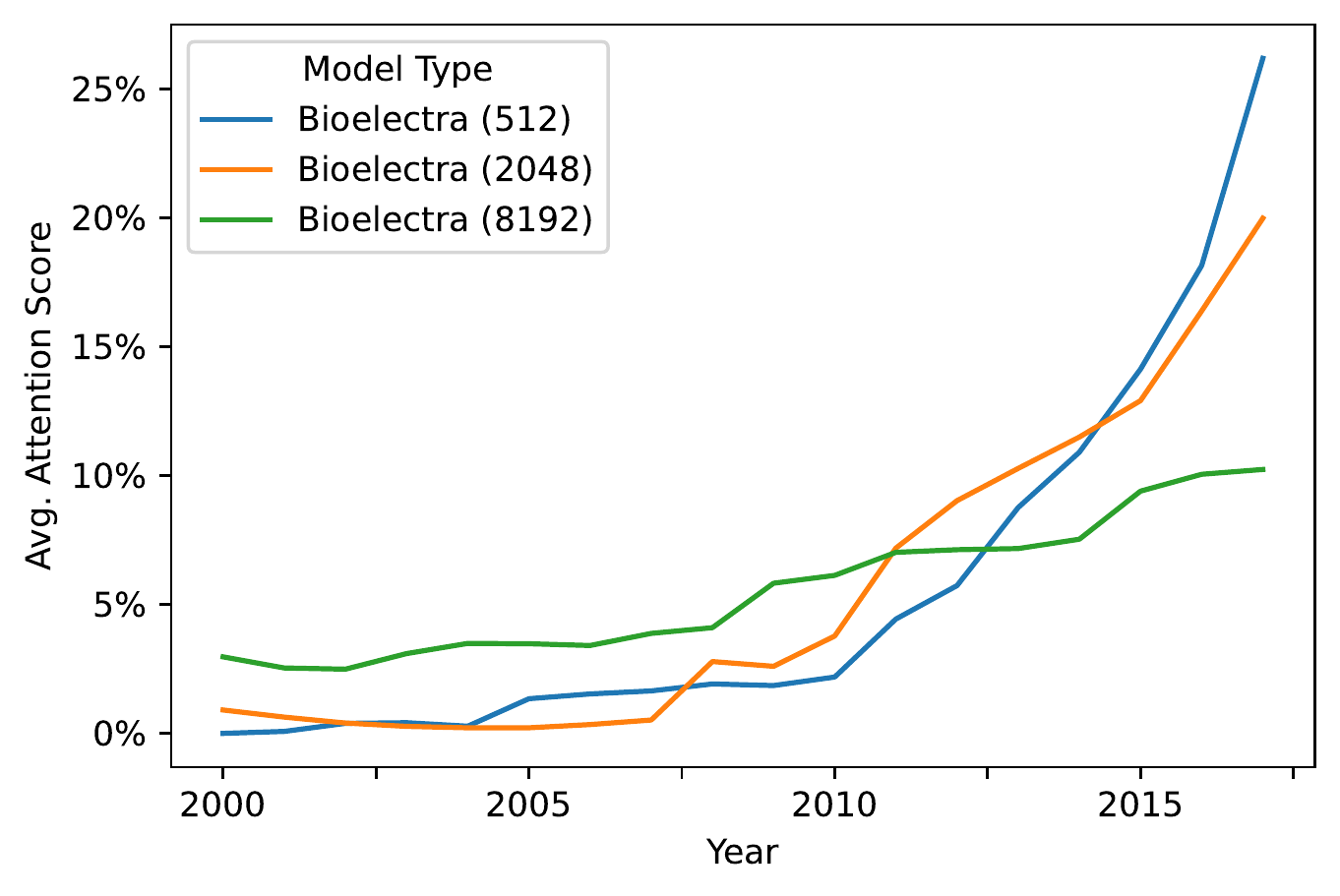} 
\end{subfigure}%
\hspace{5pt}
\begin{subfigure}{.48\textwidth}
  \centering
  \includegraphics[width=\linewidth]{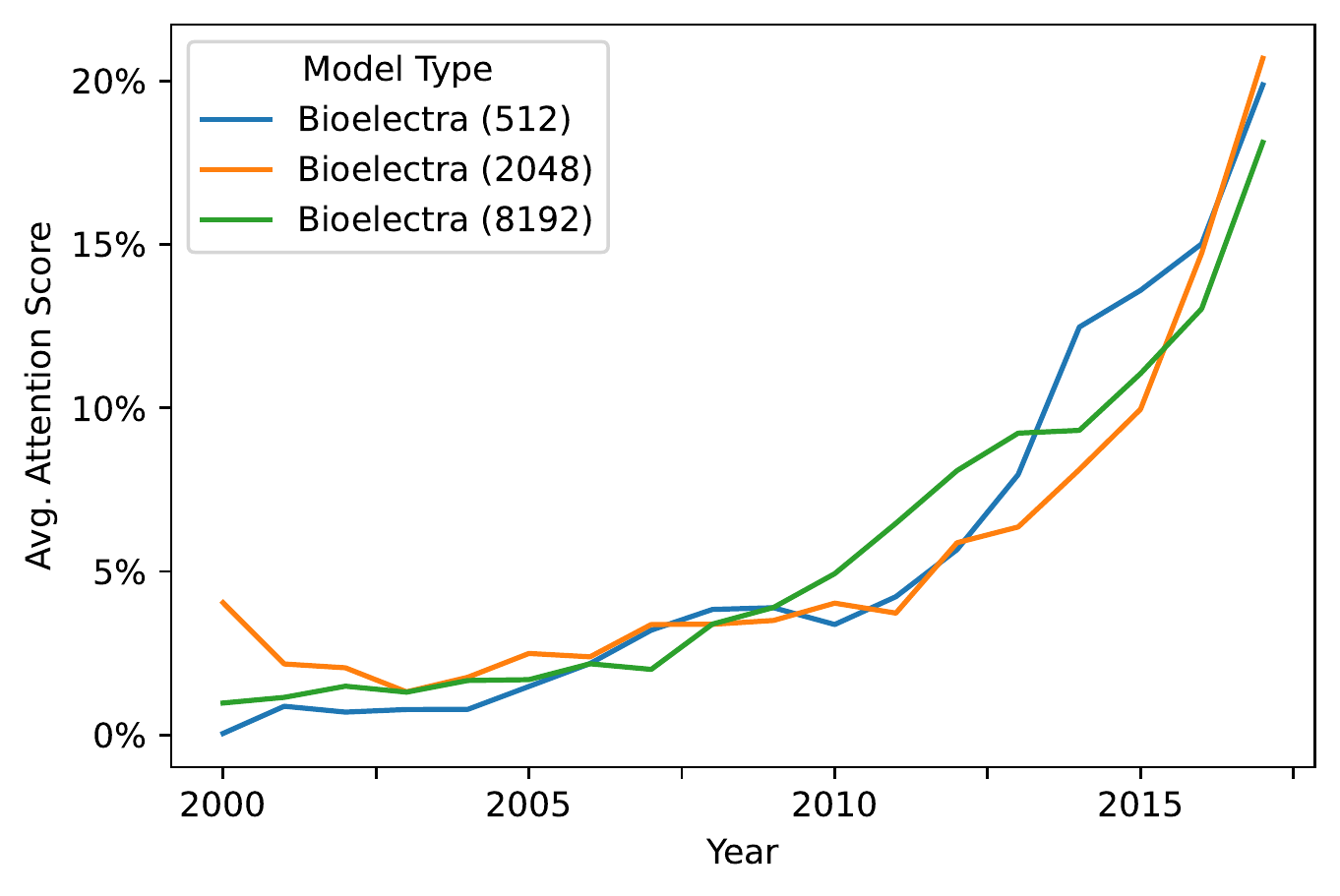} 
\end{subfigure}
\caption{\textbf{(Left)} and \textbf{(Right)} show the clinical notes \textbf{time importance} of the disease risk prediction and mortality risk prediction, respectively.
    }
\label{fig:time-importance}
\end{figure*}

\paragraph{Generalization to New Patient Data} 

We observe that there is a huge gap of performance for the baseline EHR-FFN model, especially in the diagnosis predictions of \textit{seen} and \textit{unseen} test set ($\sim$40 p.p.). 
In this case, utilizing clinical note modeling closes the performance gap on the \textit{seen} and \textit{unseen} test sets to be much narrower ($\sim$10 p.p.) on either label, especially for the BioELECTRA model with longer context length. 
This suggests that longer clinical notes interval not only improves the performance of the model on the similar patient record distribution, but also improves the performance on the out-of-distribution patient records.

\paragraph{Optimal Cut-off Interval for Disease Risk and Mortality Risk Prediction}
\label{sec:optimal}

We measure the number of clinical notes that can be processed by the models to analyze the optimal cut-off interval. Using the length statistics on our dataset, we find that our BioELECTRA (512), BioELECTRA (2048), and BioELECTRA (8192) models can encode 4, 17, 66 clinical notes on average, which correspond to the average clinical note intervals of 2-3 months, $\sim$1 year, and 3.5 years, respectively. 
As shown in Table~\ref{tab:exp-result}, for the disease risk prediction label, the utilization of longer clinical notes intervals always yields better performance, while the same trend is not observed for the mortality risk label. This evidence suggests that there are different optimal interval of clinical notes required to infer the correct prediction for different target labels.


To verify this phenomenon, we analyze the input fractions considered to be important by the models. Specifically, we retrieve 1,000 correctly-predicted samples with the highest confidence values from each of the models and collect the clinical note timestamps corresponding to the high-magnitude ($>$5\% of the total input gradient magnitude) input gradient with respect to the output prediction by using saliency map~\cite{simonyan2014saliency,yosinski2015saliency,wallace2019alleninterpret}.
The timestamps from all samples are then aggregated with yearly granularity. We denote the number of year occurrences divided by the total number of timestamps collected as \textbf{time importance} to show how likely the model attends to the clinical note from the corresponding year given the label prediction in 2018. 

As shown in Figure~\ref{fig:time-importance}, for the disease risk label, the slope of the \textbf{time importance} curves over the years become more flattened as the utilized clinical note interval widens, indicating that the \textbf{time importance} spreads more uniformly on longer clinical note intervals. Whereas for the mortality risk label, the \textbf{time importance} curve has a similar slope over different clinical notes intervals.
This evidence supports that for modeling an accurate disease risk prediction, a long clinical note interval ($\geq3.5$ years) is required. While for mortality risk prediction, a shorter clinical note interval ($\sim$2-3 months) is sufficient to reach optimal performance.

\section{Conclusion}

In this paper, we show the importance of capturing longer clinical notes for biomedical and clinical large pre-trained LMs on 6 clinical NLP tasks on the United States and Hong Kong clinical note data. Our result suggests that utilizing longer clinical notes can significantly increase the performance of LMs by $\sim$5-10\% F1-score without the loss of generalization to the unseen data. We also observe that incorporating a longer interval of clinical notes does not always entail performance improvement and there is an optimal cut-off interval depending on the target variable. Based on our analysis, we conclude that an interval of $\sim$2-3 months is the optimal cut-off for mortality risk prediction, while 3.5 years or an even longer interval of clinical notes is required to achieve the optimal performance for disease risk prediction. Future work in long-range clinical note modeling would open up opportunities towards a general solution in clinical NLP.





\section{Limitation}
Although there are many linear attention mechanisms that have been proposed~\cite{dai2019trfxl, kitaev2020reformer, beltagy2020longformer, zaheer2020big},  the exploration of linear attention in our experiments is currently limited to Longformer~\cite{beltagy2020longformer}. Furthermore, the constructed longitudinal clinical note dataset from the Hong Kong Hospital Authority EHR system cannot be made public due to the data-sharing policy. Lastly, due to the limited computational power, we only conduct the long-range clinical notes experiment for bio-lm and BioELECTRA for the n2c2 experiment and BioELECTRA for the Hong Kong longitudinal dataset. We conjecture that the performance of the long-range versions of other pre-trained models will follow similar trends to the result on existing biomedical and clinical benchmarks.

\section*{Acknowledgements}

This work has been partially funded by School of Engineering PhD Fellowship Award, the Hong Kong University of Science and Technology and PF20-43679 Hong Kong PhD Fellowship Scheme, Research Grant Council, Hong Kong.



\bibliography{emnlp2022}
\bibliographystyle{emnlp2022}

\newpage
\appendix

\section{Label Distribution}
\label{app:label-dist}
\begin{figure*}[!t]
	\centering
    \resizebox{0.88\linewidth}{!}{
        \begin{minipage}{.6\linewidth}
            \centering
            \begingroup
            \includegraphics[width=\linewidth]{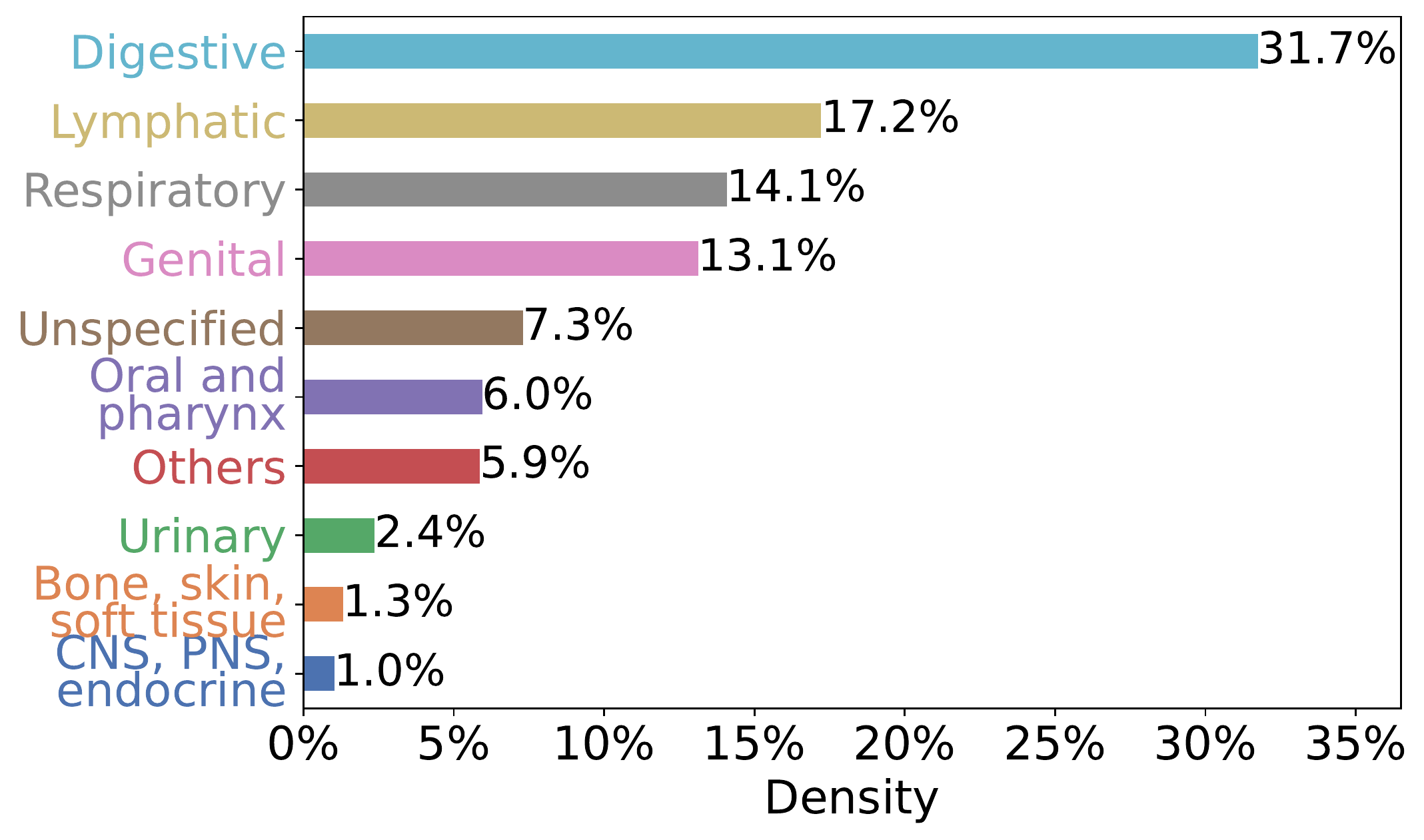}
            \endgroup
        \end{minipage}%
        \hspace{6pt}
        \begin{minipage}{.4\linewidth}
            \centering
            \begingroup
            \includegraphics[width=\linewidth]{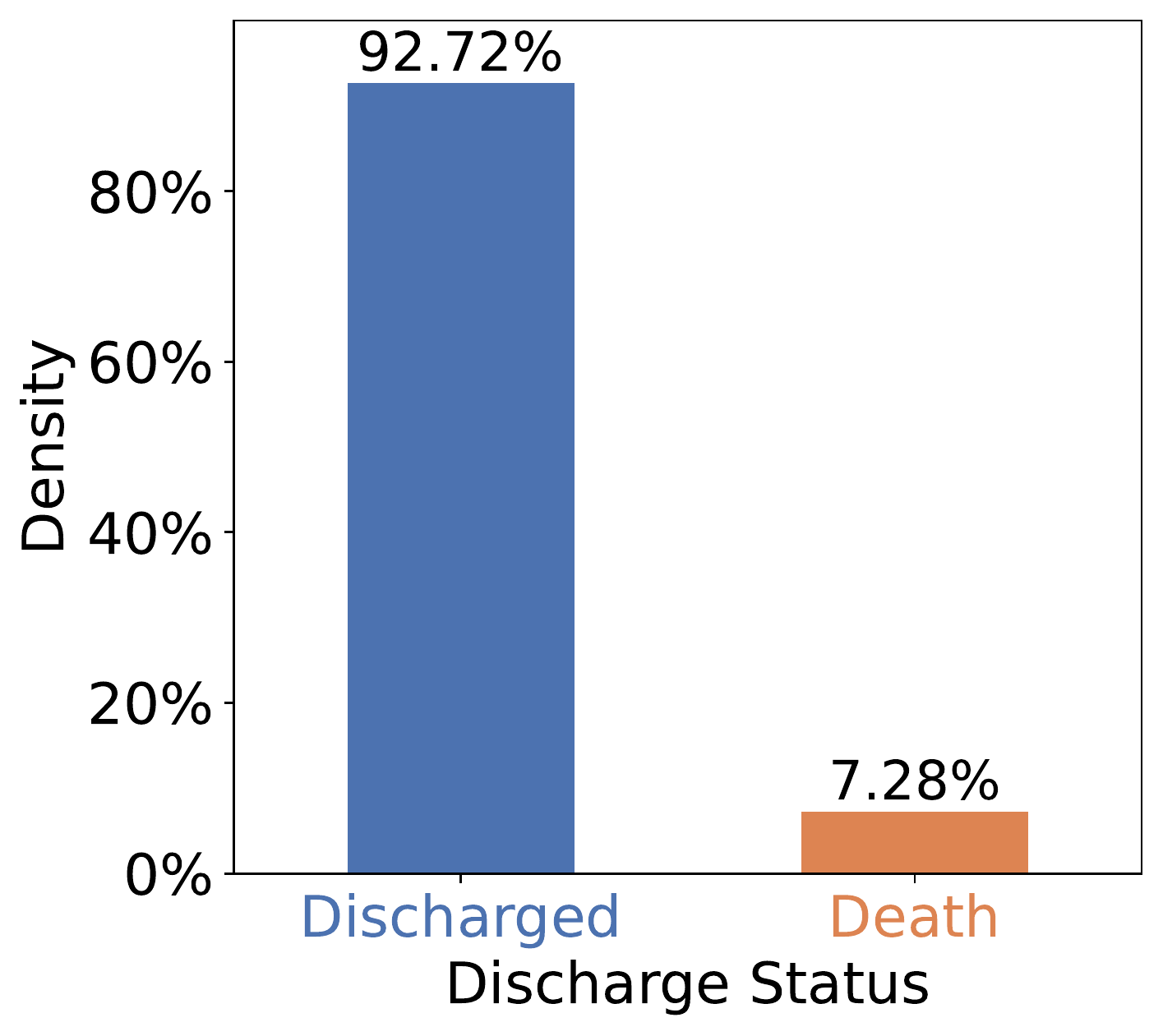}
            \endgroup
        \end{minipage}%
    }
	\caption{Label statistics of our dataset. \textbf{(Left)} shows the aggregated distribution of diagnosis based on the cancer ICD-10's site grouping\footnote{\url{https://www.icd10data.com/ICD10CM/Codes/C00-D49}}.  \texttt{Unspecified} denotes all cancer diagnoses with unspecified site. \texttt{Others} denotes diseases other than cancer. 
    \textbf{(Right)} shows the distribution of the discharge status (discharged/death) gathered from all inpatient records, which is used to define the mortality label.}
	\label{fig:label-dist}
\end{figure*}

Our Hong Kong longitudinal clinical notes dataset is extracted from Hong Kong Hospital Authority EHR system which covers records from 43 hospitals in Hong Kong. For the diagnosis, to reduce the dimensionality, we group the diagnosis labels into 79 classes. For cancer diseases, we group the diagnosis based on the cancer site categorization from the Hong Kong Cancer Registry\footnote{\url{https://www3.ha.org.hk/cancereg/allages.asp}}. While for other diseases, we take the first three digits of the ICD-10 codes as the grouping. We show the label distribution of our Hong Kong longitudinal clinical notes dataset in Figure~\ref{fig:label-dist}.

\section{Detail of EHR-FFN Model}
\label{app:tabular-models}

We derive 3,942 features from the tabular data for each encounter. We derive these features from 4 data tables: diagnosis, procedure, prescription, and inpatient data. Specifically, we generate one-hot representations for each derived feature and concatenate all the one-hot representation into a single vector . The detail of each one-hot feature is shown in Table~\ref{tab:tab-feats}. We extract the feature vectors per patient encounter. To aggregate all the historical tabular feature vectors, we aggregate the vectors into a single feature vector by summing up all the vectors producing a single high-level feature vector per patient. To learn the high-level feature vector, we employ a feed forward network with 3 hidden layers with a total size of $\sim$5M parameters. The hyperparameters of the feed forward model is shown in Table~\ref{tab:tabular-hyperparams}.

\begin{table}[!ht]
    \centering
    \resizebox{1.0\linewidth}{!}{
        \begin{tabular}{lcp{0.6\linewidth}}
            \toprule
                \textbf{Feature Name} & \textbf{Length} & \textbf{Description} \\
            \midrule
                Diagnosis Type & 1699 & Diagnosis type based on ICD-10 code\\
                Procedure Type & 127 & Procedure type based on ICD-9 code\\
                Prescription Type & 1271 & Type of presribed drug based on regional standard\\
                Prescription BNF & 73 & Type of presribed drug based on BNF Therapeutic Classification\\
                Emergency Indicator & 1 & Indicator for emergency unit admission\\
                Length of Stay & 5 & Length of stay in the hospital\\
                Age Group & 5 & Age of the patient during admission to the hospital\\
                Ward Type & 4 & Type of hospital ward\\
                Ward Sub-Care Type & 6 & Sub-type of  hospital ward \\
            \bottomrule
        \end{tabular}
    }
    \caption{Details of the tabular features}
    \label{tab:tab-feats}
\end{table}

\begin{table}[!h]
    \centering
    \resizebox{0.9\linewidth}{!}{
        \begin{tabular}{lc}
        \toprule
        \textbf{Hyperparameter settings} & \multicolumn{1}{c}{\textbf{Value}} \\
        \midrule
        \textbf{Tabular Encoder} &  \\
        \#hidden layers & 3 \\
        hidden size & {[}1024, 512, 256{]} \\
        input size & 3942 \\
        layer activation & ReLU \\
        drop out & 0.1 \\
        \bottomrule
\end{tabular}
}
\caption{Details of the model hyperparameters}
\label{tab:tabular-hyperparams}
\end{table}

\end{document}